\documentclass[letterpaper, 10 pt, conference]{ieeeconf}  

\IEEEoverridecommandlockouts                              %
\usepackage{times} 
\usepackage{amsmath} 
\usepackage{amssymb}  

\usepackage{graphicx}

\usepackage{booktabs}
\usepackage{cite}

\usepackage{algorithmic}
\usepackage{algorithm}

\title{\LARGE \bf
World-Model-Based Control for Industrial box-packing \\of Multiple Objects using NewtonianVAE
}

\author{Yusuke Kato$^{1*}$, Ryo Okumura$^{1}$ and Tadahiro Taniguchi$^{1,2}$
\thanks{$^{1}$Yusuke Kato, Ryo Okumura, and Tadahiro Taniguchi are with Digital\&AI Technology Center, Technology Division, Panasonic Holdings Corporation, Japan.}%
\thanks{$^{2}$Tadahiro Taniguchi is also with Ritsumeikan University, College of Information Science and Engineering, Japan.}%
\thanks{$^{*}$Corresponding author: \textbf{kato.yusuke001@jp.panasonic.com}}%
}

\begin{document}
\maketitle
\thispagestyle{empty}
\pagestyle{empty}

\begin{abstract}
The process of industrial box-packing, which involves the accurate placement of multiple objects, requires high-accuracy positioning and sequential actions. When a robot is tasked with placing an object at a specific location with high accuracy, it is important not only to have information about the location of the object to be placed, but also the posture of the object grasped by the robotic hand. Often, industrial box-packing requires the sequential placement of identically shaped objects into a single box. The robot's action should be determined by the same learned model.
In factories, new kinds of products often appear and there is a need for a model that can easily adapt to them. Therefore, it should be easy to collect data to train the model.
In this study, we designed a robotic system to automate real-world industrial tasks, employing a vision-based learning control model. We propose in-hand-view-sensitive Newtonian variational autoencoder (ihVS-NVAE), which employs an RGB camera to obtain in-hand postures of objects. We demonstrate that our model, trained for a single object-placement task, can handle sequential tasks without additional training. To evaluate efficacy of the proposed model, we employed a real robot to perform sequential industrial box-packing of multiple objects. Results showed that the proposed model achieved a 100\% success rate in industrial box-packing tasks, thereby outperforming the state-of-the-art and conventional approaches, underscoring its superior effectiveness and potential in industrial tasks.
\end{abstract}

\begin{figure}[t]
\centering
     \includegraphics[width=1 \linewidth]{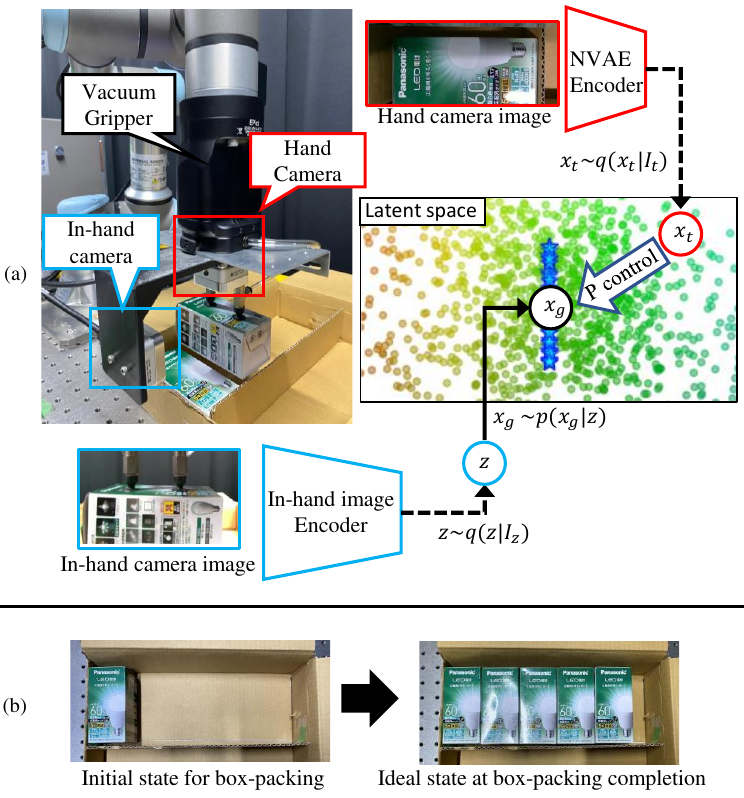}
    \caption{(a) Overview of our robotic system used in industrial box-packing. The robot is equipped with a vacuum gripper to pick up objects. Two RGB cameras are attached to the robotic hand. One obtains images from the box inside, named as ``hand camera'' and the other obtains images of the picked object posture, named as ``in-hand camera''. Latent states $x_t$ and $z$ are inferred from hand camera and in-hand camera images, respectively. An insertion position $x_g$ is generated from $z$. A vacuum gripper is positioned in the placing position by proportional control in the latent space, even with vacuum pose variations. (b) Industrial box-packing of multiple objects is our target task in this paper. This task is well known in real manufacturing plants. The initial box-packing state is shown on the left image.  The robotic system developed to perform the box-packing task of multiple objects is shown on the right image.}
\label{fig:robot_system}
\end{figure}

\section{INTRODUCTION}
Box-packing involving multiple objects is one of the most popular industrial tasks in real manufacturing plants. In most manufacturing plants, the box-packing process is performed to serve the shipment of finished products. Additionally, during a manufacturing process, the box-packing task is performed to transfer of manufactured components between various stages. As such, this task is integral to the automation of industrial operations. Automating box-packing can minimize worker fatigue and decrease labor costs. This study is focused on the industrial box-packing of multiple objects, proposing a robotic system to automate the task performed by robots equipped with a vacuum gripper.

In industrial box-packing, the standard practice aims to consolidate objects of identical shape into a box, requiring high-accuracy positioning and sequential actions (Fig. \ref{fig:robot_system}(b)). To place multiple objects into a box, the exact locations for placement must be predetermined to avoid collisions with the box walls or objects already placed in the box. Once the placing location of an object is determined, the object can be placed into the box by lowering straight down the product inventory from above, assuming the box is positioned on a horizontal table. Therefore, to avoid bottlenecks during box-packing, high-accuracy positioning is required. Moreover, sequential controls executed by the robot are critical for the box-packing task when multiple objects are involved. The placement of each subsequent item relies on knowledge of the position of the previously placed object.
A key challenge in deploying robot systems in factories is their adaptability to new kinds of products. As new products frequently emerge in a factory setting, the robotic system must be flexible enough to manage them. Traditional rule-based approaches demand rule modifications to accommodate new objects, usually requiring expert intervention.With the trend of high-variety, small-batch production, system integrators bear a considerable burden when handling new products, which can hinder the introduction of robotic systems into factories. Therefore, even non-expert factory workers should be able to collect data for training and train models for new products.

Robotic systems that execute dense packing have been investigated in previous studies. For instance, a study \cite{wang} demonstrated the packing of geometrically complex non-convex objects by determining the optimal object positioning and orientation. Dong proposed a box-packing strategy for placing objects into a box \cite{dong} that prevented collisions with environmental objects using tactile sensors. These studies dealt with various object shapes in warehouse setting; however, in industrial box-packing scenarios, dense box-packing of objects is typically required. Therefore, high-accuracy positioning is essential. Furthermore, the industrial box-packing task requires the placing of multiple objects in a sequential procedure.

Okumura et al. developed a tactile-sensitive Newtonian variational autoencoder (TS-NVAE)\cite{ts-nvae} for a task requiring incredibly high-accuracy positioning, such as inserting a USB connector. TS-NVAE is based on the NewtonicanVAE (NVAE) \cite{nvae}, which is a world model \cite{world_model} type. The latent transition model was trained with Newton's equations and a goal state prediction model was jointly trained using tactile sensor information. This enabled the prediction of the goal position for the connector insertion from the in-hand object pose obtained from the tactile sensor, which was attached to the inside of the robotic finger. Given that the USB connector was positioned directly above the socket, it could be inserted by lowering it in a straight path.
The benefit of this approach is the ease of data collection for model training. The robot randomly walk over the target object to collect training data, capturing a sequence of images and corresponding motion vectors. Consequently, training a new model to insert USB sockets with different appearances can be as easily as a factory worker performing the data collection, eliminating the need for expert intervention.
Although TS-NVAE is a powerful model in visual feedback control, it poses limited applicability to general industrial tasks. This is primarily because of the difficulty of integrating tactile sensors with vacuum grippers, typically used to pick up objects in industrial tasks. Box-packing tasks are usually performed by robots equipped with vacuum grippers. These robots require high-accuracy positioning to place objects into a box in the same orientation and alignment. 
Furthermore, unlike connector insertion, box-packing requires repetitive actions to fill the box.

In this study, we propose an innovative approach for automating the box-packing task performed by robots, the in-hand-view-sensitive NewtonianVAE (ihVS-NVAE). Unlike the tactile-sensor information utilized in TS-NVAE, our method employs in-hand-view sensor data. The conventional use of vaccum grippers in industrial tasks creates difficulties in integrating tactile sensors.
To this end, we employ standard RGB cameras as in-hand-view sensors and generalize a goal state prediction model using standard RGB camera data. The RGB camera is attached to the robotic hand to capture images of the objects posture during vacuuming. Furthermore, we demonstrate that our method can be used in a sequential task without any additional training; it generalizes the positioning method from observation images based on the world model. The box-packing task requires similar actions to be repeated before completion, considering the conditions inside the box. However, it is difficult to train visual feedback-control models for all possible state variations within the box.

The major contributions of this research are stated as follows:
\begin{itemize}
\item We generalize a goal state prediction method using an RGB camera to acquire information on the posture of vacuumed objects.
This data collection method for unseen objects does not require expertise.

\item We show that our proposed positioning method can be applied to sequential tasks without the additional training to account for condition variations. Repetitive actions can be performed until the task is complete.
To achieve this, we take advantage of VAE's property that values of close latent variables are estimated for similar inputs.

\item We conduct evaluation experiments for the industrial box-packing task involving multiple objects using actual robots. The proposed method outperformed state-of-the-art (SOTA) methods, which do not require expertise in training data collection.
\end{itemize}

The superior performance of the proposed method is demonstrated in the industrial box-packing process. It achieves more accurate positioning than SOTA methods for two types of objects and can sequentially placing more objects into a box than those placed using SOTA methods.

\section{PRELIMINARY CONCEPTS}
\subsection{NewtonianVAE}
A world model \cite{world_model} is a generic term  for models incorporating state inference and transition in partially observable Markov decision processes. The joint optimization of inference and transition models leads to accurate and sample-efficient training. Four types of control methodologies used trained world models; they include policy \cite{lee, hafner, levine, dreaming, dreamingv2}, model predictive control \cite{hafner2, okada, daac}, optimal control \cite{levine2, levine3, watter}, and proportional control. The NewtonianVAE (NVAE)\cite{nvae} belongs to the world model category and enables proportional control in a latent space. The latent variables of NVAE was modeled to follow Newton's equations, as indicated below:
\begin{eqnarray}
\frac{d\mathbf{x}}{dt}&=&\mathbf{v} \\
\frac{d\mathbf{v}}{dt}&=&A(\mathbf{x},\mathbf{v})\cdot \mathbf{x} + B(\mathbf{x},\mathbf{v})\cdot \mathbf{v} + C(\mathbf{x},\mathbf{v})\cdot \mathbf{u}
\end{eqnarray}
The model was trained to maximize the lower bound given below:
\begin{align}
\mathcal{L}=&\mathbb{E}_{q(\mathbf{x}_t\vert\mathbf{I}_t)q(\mathbf{x}_{t-1}\vert\mathbf{I}_{t-1})}[\mathbb{E}_{p(\mathbf{x}_{t+1}\vert\mathbf{x}_t,\mathbf{u}_t;\mathbf{v}_{t+1})}\rm{log}\mathit{p}(\mathbf{I}_{t+1}\vert\mathbf{x}_{t+1})\notag \\
&-KL(q(\mathbf{x}_{t+1}\vert\mathbf{I}_{t+1})\Vert\,p(\mathbf{x}_{t+1}\vert\mathbf{x}_t,\mathbf{u}_t;\mathbf{v}_{t+1}))] 
\end{align}

where $\mathbf{x}_t\in\mathbb{R}^D$ represents the latent state that is inferred by a posterior inference model, $\mathbf{x}_t\sim q(\mathbf{x}_t\vert\mathbf{I}_t)$; $\mathbf{u}_t\in\mathbb{R}^D$ represents an action, and $\mathbf{I}_t$ denotes an observation image.
$\mathbf{v}_t$ is calculated as $\mathbf{v}_t=(\mathbf{x}_{t}-\mathbf{x}_{t-1})/\Delta\,t$.

In this model, the transition prior can be expressed as
\begin{align}
p(\mathbf{x}_{t+1}\vert\mathbf{x}_t,\mathbf{u}_t;\mathbf{v}_{t+1})=\mathcal{N}(\mathbf{x}_{t+1}\vert\mathbf{x}_t+\Delta\,t\cdot\mathbf{v}_t,\,\sigma^2) &\\
\mathbf{v}_{t+1}=\mathbf{v}_t+\Delta\,t\cdot(A\mathbf{x}_t+B\mathbf{v}_t+C\mathbf{u}_t) &
\end{align}
with
\begin{align}
    [A,\:\rm{log}(\mathit{-B}),\: \rm{log}\mathit{C}]=\rm{diag}(\,\mathit{f}(\mathbf{x}_t,\mathbf{v}_{t},\mathbf{u}_t))
\end{align}
where $f$ is a neural network with a linear output activation. Additional detailed descriptions can found in \cite{nvae}.

\subsection{Goal State Prediction}
The TS-NVAE\cite{ts-nvae}, developed by Okumura et al., can predict a goal-state using tactile-sensor. They jointly trained a goal state prediction model using tactile sensor images along with the NVAE latent model. They collected data of image-action sequences $D_x=\{(\mathbf{I}_1,\mathbf{u}_1), ..., (\mathbf{I}_T,\mathbf{u}_T) \}$ and a pair of image data $D_z=(\mathbf{I}_z, \mathbf{I}_g)$, where $\mathbf{I}_t$ is a camera image, $\mathbf{I}_z$ is a tactile sensor image, and $\mathbf{I}_g$ is a camera image at the goal position. The loss function is represented as $\mathcal{L}_x+\mathcal{L}_z$ such that:
\begin{align}
\mathcal{L}_x =& \: \mathbb{E}_{q(\mathbf{x}_t\vert\mathbf{I}_t)q(\mathbf{x}_{t-1}\vert\mathbf{I}_{t-1})}[-\text{log}p(\mathbf{I}_{t+1}\vert\mathbf{x}_{t+1})\notag \\
&+KL(q(\mathbf{x}_{t+1}\vert\mathbf{I}_{t+1})\Vert\,p(\mathbf{x}_{t+1}\vert\mathbf{x}_t,\mathbf{u}_t))] \\ \notag\\
\mathcal{L}_z =& \: \mathbb{E}_{q(\mathbf{z}\vert\mathbf{I}_z)}\mathbb{E}_{p(\mathbf{x}_g\vert\mathbf{z})}[-\text{log}p(\mathbf{I}_z\vert\mathbf{z}) \notag\\
&-\text{log}p(\mathbf{I}_g\vert\mathbf{x}_g) + KL(q(\mathbf{x}_g\vert\mathbf{I}_g)\Vert\,p(\mathbf{x}_g\vert\mathbf{z}))] 
\end{align}
To simplify the transition model, they used the transition prior given below:
\begin{align}
p(\mathbf{x}_{t+1}\vert\mathbf{x}_t,\mathbf{u}_t)=\mathcal{N}(\mathbf{x}_{t+1}\vert\mathbf{x}_t+\Delta\,t\cdot\mathbf{u}_t,\,\sigma^{2}_{x}) 
\end{align}
To achieve their goals, Okumura et al. utilized the built-in Cartesian velocity control of a robot arm, where $\mathbf{u}_t$ represents the reference velocity in Cartesian space. Furthermore, they introduced two domain knowledge components into the latent distribution. Initially, they established that the transition uncertainty of robots, $\sigma_x$ aligns with the nominal repeated positioning accuracy of real robots. Subsequently, they introduced the Kullback–Leibler (KL) terms to regularize $\sigma_x$ with respect to the position variation $\sigma^{2}_{g}$ as detailed below:
\begin{align}
\mathbb{E}_{q(\mathbf{z}\vert\mathbf{I}_z)}[KL(q(\mathbf{x}_g\vert\mathbf{I}_g)\,\Vert\,\mathcal{N}(0, \sigma^{2}_{g})) \notag\\
+KL(p(\mathbf{x}_g\vert\mathbf{z})\,\Vert\,\mathcal{N}(0, \sigma^{2}_{g}))]
\end{align}
These operations ensure that the latent space shares the same axes and scales as the Cartesian coordinate system in the physical space.

\section{In-hand-View-Sensitive Newtonian VAE}
\subsection{Proximity Camera for In-hand Object Posture}
In the proposed method, we utilize in-hand-view sensors to predict a goal state for the box-packing task. To generalize the applicability of our method, we attached a standard RGB camera as an in-hand-view sensor to the robotic hand to capture proximity images of the in-hand object posture. Although tactile sensors attached to a 2-finger gripper can easily obtain an accurate posture of the in-hand object, their usage is limited in certain scenarios. For instance, when a vacuum gripper is employed to pick up an object, it is difficult to obtain the in-hand object posture using tactile sensors. RGB cameras deployed as in-hand-view sensors require some preprocessing to accurately determine the in-hand object posture. Proximity images of an in-hand-view sensor include not only information about the posture of the in-hand object but also unnecessary information such as the background environment. Therefore, the image is cropped to eliminate this unnecessary information in advance. The position of the object within the proximity images can be known and it is fixed, since the positional relationship between the robotic hand and the camera is predetermined.

\subsection{Utilization of Latent Space Characteristics}
\begin{figure}[t]
\centering
     \includegraphics[width=1 \linewidth]{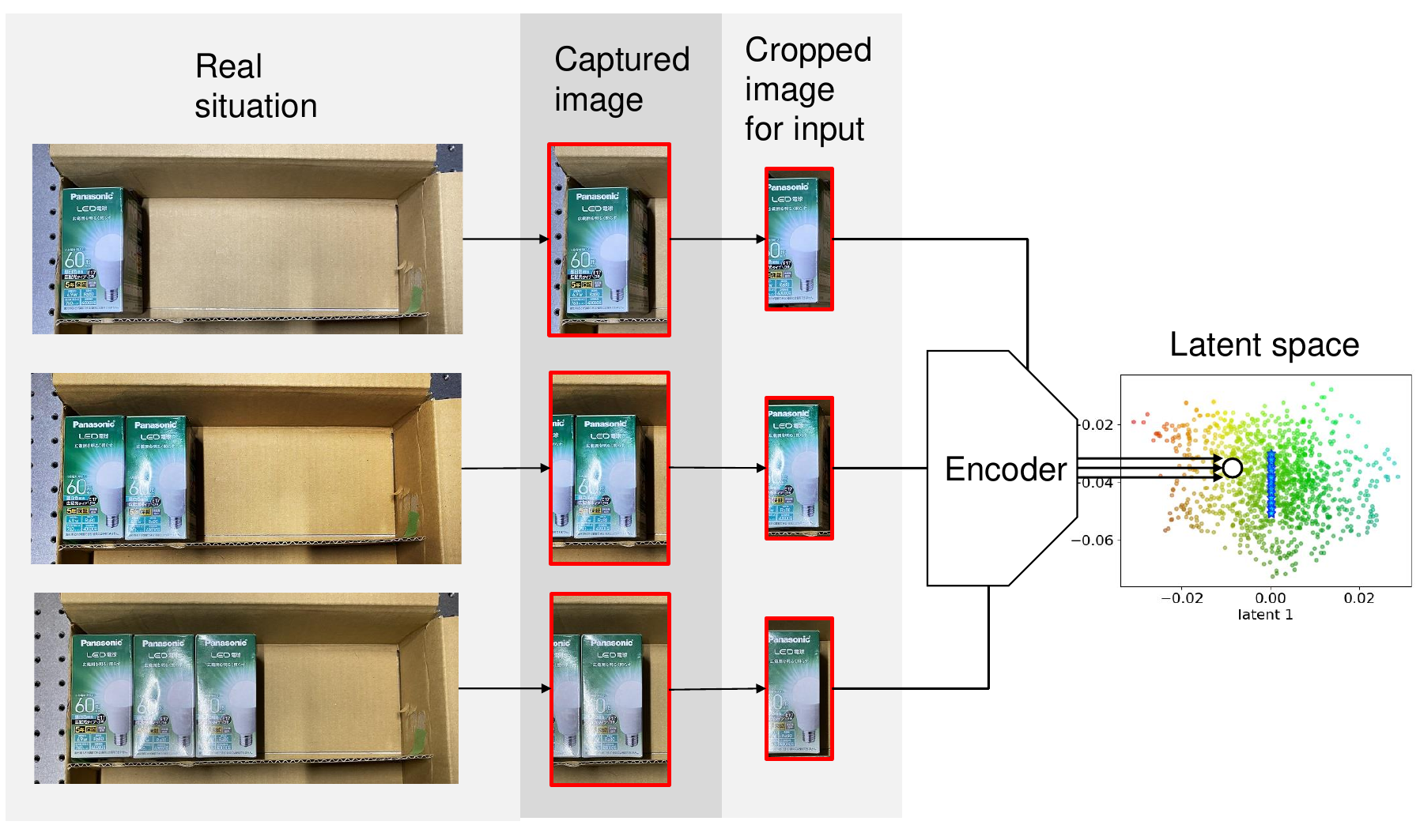}
    \caption{In the VAE encoder, if the input images exhibit a similar appearance, the estimated latent variables will also take similar values. Thus, the captured images were cropped such that the model could treat them in the same state, even if they were actually in different states.}
\label{fig:crop}
\end{figure}

Our ihVS-NVAE model estimates latent variables from input images. In a model utilizing VAE, when two input images have the same appearance, their latent variable values are closely aligned. We leverage this characteristic to achieve repeated task execution, which will be introduced later. During the model training, an image subjected to specific cropping is used as input.
\begin{align}
I^{input}_t = crop(I^{captured}_t)
\end{align}
By applying cropping to the captured image, the states under different circumstances can be treated as the same state (Fig. \ref{fig:crop}).


\subsection{Sequential Industrial Box-packing of Multiple objects}
\label{IndustrialPacking}
\begin{figure}[tpb]
\centering
     \includegraphics[width=1. \linewidth]{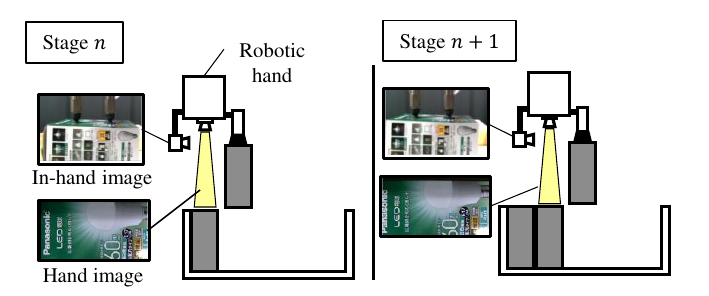}
    \caption{The world model concept can be applied to perform industrial box-packing tasks. During the data collection phase, hand and in-hand images are collected at stage $n$. After training, the model executes the box-packing task at stage $n$. The model also executes the task at stage $n+1$ because the environment is observed locally, and the observed images are similar at each stage.}
\label{fig:sequntial_input}
\end{figure}

The industrial box-packing of multiple objects can be processed as a sequential task. The position for object placement can be predicted based on the placement of the object during the previous stage (Fig. \ref{fig:sequntial_input}). The models based on NVAE, including the proposed ihVS-NVAE, are trained with the relative positions between the robots and the object from the observation images. Consequently, the proposed method can be applied to the sequential positioning in industrial box-packing without any additional training by locally observing the previously placed object. This is attributable to the property of VAE \cite{vae} that maps similar observation images in close proximity within the latent space.

In this study, the sequential industrial box-packing of multiple objects using our method is executed following Algorithm \ref{alg1}. To compute the action of robot $a$, we implemented a simple proportional control as the $controller()$.

We assume that one or more objects are already in the box. 
In industrial environments, packing boxes are typically prepared by workers. In such instances, this problem is readily resolved if they place the initial object in the box. Therefore, our algorithm is capable of being implemented in an industrial environments.

\begin{algorithm}[tpb]
    \caption{Industrial box-packing of multiple objects}
    \begin{algorithmic}
    \label{alg1}
    \REQUIRE Number of Stages $\mathit{N}$, episode length $T$
    \REQUIRE The position $p_{pick}$ to pick up an object
    \REQUIRE The start position $p_0$ to move above a box
    \REQUIRE $p(\mathbf{x}_g\vert\,\mathbf{z}), q(\mathbf{z}\vert\,\mathbf{I}_z)$, controller()
    \FOR{$n=1...N$}
    \STATE Pick up a object at $p_{pick}$
    \STATE Move to $p_0$ and obtain an image $\mathbf{I}_z$
    \STATE $\mathbf{z} \sim q(\mathbf{z}\vert\,\mathbf{I}_z)$
    \STATE $\mathbf{x}_g \sim p(\mathbf{x}_g\vert\,\mathbf{z})$
        \FOR{$t=1...T$}
            \STATE Obtain an image $\mathbf{I}_t$ at current position $p_t$
            \STATE $\mathbf{x}_t \sim q(\mathbf{x}_t\vert\,\mathbf{I}_t)$
            \STATE  $a \gets \text{controller}(\mathbf{x}_t, \mathbf{x}_g)$
            \STATE Execute action $a$
        \ENDFOR
    \STATE Place the object at $p_T$
    \STATE $p_0 \gets p_T$
    \ENDFOR
    \end{algorithmic}
\end{algorithm}

\section{Experiment}
\subsection{Hardware Configuration}
\begin{figure}[tpb]
\centering
     \includegraphics[width=1. \linewidth]{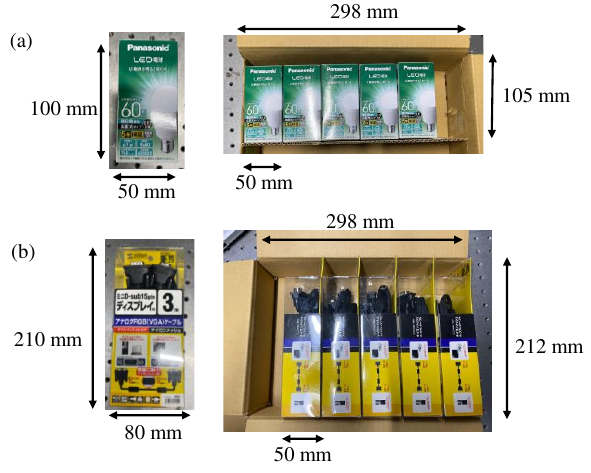}
    \caption{Objects used in the box-packing experiment. (a) LED package with size 50 mm $\times$ 100 mm $\times$ 50 mm. (b) Cable package with size 80 mm $\times$ 210 mm $\times$ 50 mm. Each box can store up to five objects. Compared to Cable packages, LED packages are smaller and have a wider margin within the box. }
\label{fig:objects}
\end{figure}

Our experimental robotic system is shown in Fig. \ref{fig:robot_system}(a). We employed a Universal Robots UR3e as a 6-DOF robotic arm. The end effector was a ROBOTIQ EPick, which is a vacuum gripper. The gripper was equipped with two vacuum pads capable of vacuuming at two points. Two RGB cameras were attached to a tool mount at the end effector. Both cameras were the Intel RealSense D405. Although these cameras can capture depth images, we did not use this function. One camera was used to capture images of the box inside, and the other was used to capture images of the vacuum position. Both cameras and the vacuum pads were attached to the gripper. The relative positions of the camera and the suction pad attached to the robotic hand are shown in Fig. \ref{fig:robot_system} (a). In all experiments, we used the built-in Cartesian velocity control of the UR3e robotic arm system for transitions in the X and Y axes. As shown in Fig \ref{fig:objects}, we prepared two differently-sized objects for our experiment. In this study, these two objects are named ``LED package'' (Fig. \ref{fig:objects}(a)) and ``Cable package'' (Fig. \ref{fig:objects}(b)). Each box can accommodate up to five objects. LED packages, being smaller, have a wider margin within their box than Cable packages. Consequently, Cable packages require a more accurate positioning than LED packages to be placed into the box.

\subsection{Data Collection}
\begin{figure}[tpb]
\centering
     \includegraphics[width=1. \linewidth]{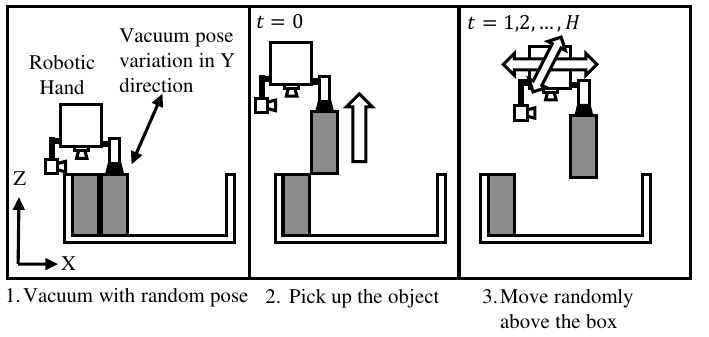}
    \caption{Data collection process. Initially, the robot vacuums an object with a random pose. Thereafter, it picks up the object in an upward direction. At this instant, $\mathbf{I}_z$ and $\mathbf{I}_0$ are observed. $\mathbf{I}_0$ is treated as $\mathbf{I}_g$ during the training of the placing-position estimator $p(\mathbf{x}_g\vert\mathbf{z})$. Subsequently, the robot moves randomly above a box to collect the transition data at each time step $t$; time horizon \textit{H}: 20, number of episodes: 60.}
\label{fig:data_collection}
\end{figure}

This section detailes the process of collecting data to train the ihVS-NVAE model. To train the model, we collected robot actions and observation images. We present a data collection method that does not require expertise. This data collection process can be easily performed by factory workers without expertise.
Fig. \ref{fig:data_collection} shows the proposed data collection process. The box location and the object vacuuming position were considered during the data collection. Initially, one object was picked up from the box. The vacuum position was randomly changed within a range of $\pm0.01$ [m] per episode in the longitudinal direction of the package. This is intended to enhance the generalization capability in response to alterations in the vacuum positions. Let $t=0$ indicate the time of picking up the object, and $\mathbf{I}_0$ and $\mathbf{I}_z$ represent the hand and in-hand camera images at this instant, respectively. This instantaneous position can be used as the place position in an industrial packing task ($\mathbf{I}_g = \mathbf{I}_0$).
Thereafter, pairs of the observed image $\mathbf{I}_t$ and the action $\mathbf{u}_t$ were acquired by a random move performed by the robot to observe the object in the box. The actions were randomly determined in the X- and Y-axis directions within a range of $\pm0.01 $ [m/s]. This process enables annotation-free data collection without operating the system for each episode.
The process was repeated 60 times to complete the data collection process. The control frequency was 2 [Hz] and the length per episode was $10$ [s]. Overall, 1200 transitions were collected for training.

If only the initial position where the object is vacuumed is provided by an operator, training data can be collected by randomizing the vacuum position and action. 
Since neither image annotation nor environment resetting is required, even factory workers who do not have expertise can collect the training data.


\subsection{Training}
We train individual models for each of the two objects we have prepared.
We prepared a network configuration as shown in Fig.\ref{fig:robot_system}(a). Latent states $x_t$ and $z$ were inferred from the hand camera and in-hand camera images, respectively. The dimension of the latent vector was 2, and an insertion position $x_g$ was generated from $z$. Network training was performed in accordance with \cite{ts-nvae}. The fundamental difference is in terms of the encoder configuration. Resized 64 $\times$ 64 pixel RGB images were used as inputs for both camera encoders. Image encoders were used as a simple convolutional neural network (CNN). These encoders were trained from scratch, as the input images were modified from a rectangular to a square format. The image sizes obtained from the hand and proximity cameras were 300 $\times$ 120 and 360 $\times$ 300 for the LED package and 300 $\times$ 120 and 720 $\times$ 300 for the Cable package, respectively. Given that an LED package is smaller than a Cable package, the proximity image size of the in-hand-view sensor was also reduced to avoid capturing external information. The encoder output was used to infer the posterior model through fully-connected networks. These fully-connected networks had two hidden layers with 16 units and employed the LeakyReLU \cite{leakyrelu} activation to infer $\mathbf{x}_t$ and $\mathbf{z}$. The goal position $\mathbf{x}_g$ was inferred from $\mathbf{z}$. We assumed that all prior and posterior models conformed to the Gaussian distribution. All models were trained using the Adam \cite{adam} optimizer with a learning rate of $3 \times 10^{-4}$.

\section{EVALUATION}

\subsection{Visualization of the Latent Space}
\begin{figure}[tpb]
\centering
     \includegraphics[width=1. \linewidth]{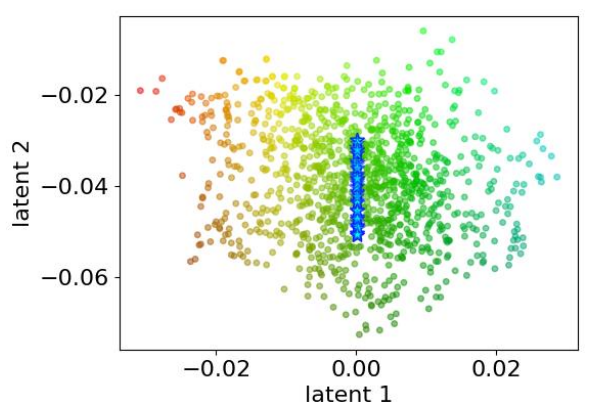}
    \caption{Visualization of latent space obtained using our trained model. Color markers represent the positions in the Cartesian coordinate system. Color variation from red to green represents the \textit{X} position; color saturation represents \textit{Y}, where high saturation indicates a high \textit{Y} position. Blue stars represent the estimated placement position obtained from the in-hand-view sensor images.}
\label{fig:latent_map}
\end{figure}

To evaluate our model, which predicts the goal states using RGB camera images of a vacuumed object posture, we visualized the latent space obtained by our trained model. Fig. \ref{fig:latent_map} presents the resultant latent space after model training. Variations in hue and saturation correspond to the \textit{X} and \textit{Y} positions in the Cartesian coordinate system, respectively. The blue star represents the location of the goal state $\mathbf{x}_g$ estimated using the in-hand image $\mathbf{I}_z$. As th random move of robot was centered on the goal state $\mathbf{x}_g$ during data collection, we observed that $\mathbf{x}_g$ can learn to be in the center of the latent space after model training. The results indicated that the goal states can be successfully estimated using in-hand images.

\begin{table}[tb]
    \centering
    \caption{Industrial packing task performance\\ (The accuracy is indicated by the mean $\pm$ standard deviation)}
    \begin{tabular}{lcc}
    \toprule
    Method & \multicolumn{2}{c}{Accuracy [mm]} \\
    \cmidrule(lr){2-3}
     & LED package & Cable package\\
    \midrule
        ihVS-NVAE (ours) & $2.22 \pm 1.27$  & $1.18 \pm 0.65$ \\
        CFIL\cite{cfil}+In-hand-viewCNN &  $6.16 \pm 3.78$ & $6.21 \pm 2.72$ \\

    \bottomrule
    \end{tabular}
    
    \label{tab:result}
\end{table}

\subsection{Positioning Performance}
\begin{figure}[tpb]
\centering
     \includegraphics[width=1. \linewidth]{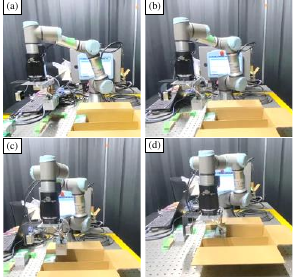}
    \caption{These pictures show a robot executing  box-packing task in the real world. (a) The robot picks up an object. (b) The robot moves to initial position and captures a proximity image to estimate the goal location. (c) The robot moves to its initial position to accomplish the task. If the estimated goal position differs from its current position, our model determines the action of the robot.(d) When the robot arrives at the estimated goal position, it is lowered straight down to complete the task.}
\label{fig:experiment}
\end{figure}

The goal location was evaluated based on the accuracy of the estimated goal states using in-hand images. The robot picked up an object in the box in the same way as in the data collection process. The position at that moment was stored as the correct position, and the robot moved to a random location. Subsequently, we calculated the position move error (relative to the correct position) through goal estimation using proximity images. For comparison, we employed the coarse-to-fine imitation learning (CFIL) method\cite{cfil}, a SOTA model for positioning. 
This method does not require expertise to collect data for training models. 
In CFIL, the action is estimated from the image of a single camera attached to the robotic hand, so the object posture within the hand is not considered. 
To counteract this issue, we added a network to the CFIL that estimates changes in object posture from in-hand camera images. This network consists of convolutional and fully-connected layers, which obtain images from the in-hand camera as input and output correction values for the pose of objects within the hand. After the execution of the CFIL is completed, correction is performed to match the object posture, making it possible to consider the posture of objects inside the hand. We refer to this approach as CFIL+in-hand-viewCNN, which we used as the baseline for comparison.
As a result, our method is more accurate than SOTA methods, with a positioning accuracy of 2.22 mm for the LED package and 1.18 mm for the Cable package.
The CFIL+in-hand-viewCNN combines a positioning method using hand camera images with a correction for object posture using in-hand camera images. As a result, the errors of each estimation accumulate. In contrast, our method integrates information from both the hand camera and in-hand camera, resulting in smaller errors. Therefore, when performing tasks that require consideration of object posture, it is important to integrate the information from hand camera and in-hand camera images. Fig. \ref{fig:experiment} shows a robot performing a box-packing task.

\subsection{Sequential Industrial Box-Packing Performance}
\begin{figure}[tpb]
\centering
     \includegraphics[width=1. \linewidth]{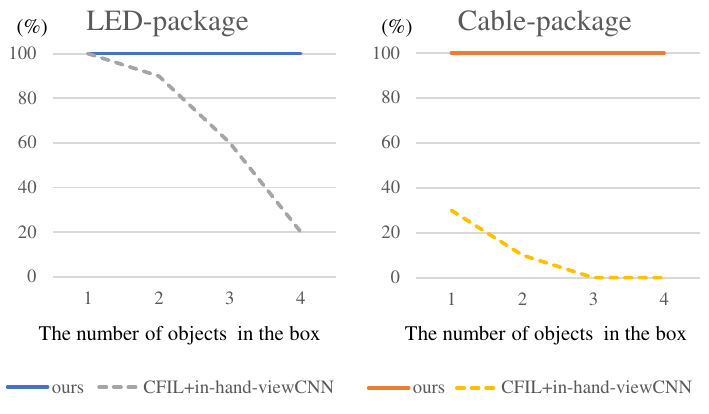}
    \caption{The success rates in industrial box-packing of multiple objects are shown. The horizontal axis denotes the number of objects packed in a box. The vertical axis shows the success rate for various types of objects using different methods. The solid and dashed lines respectively represent the results of our method and the comparative method. In our approach, all four objects (including LED packages and Cable packages) could be placed into the box, thus completing the industrial box-packing task. In contrast, with the CFIL+in-hand-viewCNN method, the success rate declined, as the sequence progressed.}
\label{fig:success_rate}
\end{figure}

We conducted sequential industrial box-packing of multiple objects with a physical robot, following the procedure detailed in Algorithm \ref{alg1}. Four objects were sequentially placed one-by-one in the box. One object was already placed in the box, setting the initial state. To evaluate the performance, we conducted sequential industrial box-packing of multiple objects up to 10 times and evaluated the success rate as the sequence progressed. The results are shown in Fig. \ref{fig:success_rate}. Our method achieved a 100\% success rate at all stages. In contrast, with the CFIL+in-hand-viewCNN method, the success rate declined as the sequence progressed. In sequential tasks, errors tend to accumulate as results previous stages are used as observations. Despite the wide margins of the LED packages, the success rate falls after the second stage due to the accumulation of positioning errors at each stage. Our approach allows for the training of the neural network with hand and in-hand camera images, facilitating the inference of current and goal states, resulting in improved accuracy.

\section{DISCUSSION}
In our robotic system, objects were picked up by vacuum grippers at two points. The selected objects possessed a balanced center of gravity, which was sufficient to vacuum grip at the object center. By changing the vacuum positions, the object orientation varied, but only along the longitudinal direction of the package. However, an object with an unbalanced center of gravity would tilt with respect to the horizontal plane when picked up, thereby hindering the efficiency of the packing process. In this experiment, we assumed that the object could be placed back in its original position if it was lowered straight down to the same position from which it was picked up from the box. However, this is not possible for a tilted object. To address this issue, it would be necessary to generate placement trajectories that consider the tilt or understand the vacuumed object posture, so that it does not tilt. The proposed method can be extended to understand the vacuumed object pose with increased accuracy using depth images. Additionally, we aim to introduce the concept of the multi-view world model\cite{multi-dreaming} in our method. We aim to generate trajectories for object placement using additional images of the object and box.

Our approach employed a proximity camera to aquire images of the in-hand object, enabling positioning that considers the vacuum position displacement. Since it is necessary to extract features of the objects appearance (not its shape as with a tactile sensor), estimating the displacement of the vacuum position using our method would be difficult if only a few features of the object surface were known. By using depth and RGB images, it is possible to recognize the object posture with numerous formation.

\section{CONCLUSIONS}
In this study, we proposed a method to automate real-world industrial tasks. Box-packing of multiple objects is a common procedure in manufacturing environments. The process requires accurate positioning and sequential actions to efficiently pack objects into a box.
In industrial settings, new kinds of products are often introduced. With conventional rule-based methods, it is required expertise to redesign models each time to adapt these new products. This presents a barrier to the wide spread introduction of robotic systems in factories.
To address this challenge, we proposed a generalization of the TS-NVAE model, using proximity images of the in-hand object posture and proposed the ihVS-NVAE.
We employed RGB cameras to obtain the posture of the in-hand object. 
Our method enables factory workers, even those without expertise, to collect the data required for model training. The incorporation of image cropping into the input data for the model enables the execution repetitive tasks by allowing the model to  treat different states as the same state, thereby simplifying task completion.
The proposed approach achieved more accurate positioning than SOTA methods. Furthermore, this method is applicable to sequential tasks, which form a key requirement in industrial box-packing scenarios. Using world model property, our model is capable of executing tasks in a sequence without any additional training for each separate stage of the process. The results showed that our method could place more objects in a box sequentially than those placed by SOTA methods. This is highlights the potential of our approach in contributing remarkably to the automation of industrial task.

\section{ACKNOWLEDGEMENT}
This work was partially supported by JST Moonshot R\&D, Grant Number JPMJMS2033.

\addtolength{\textheight}{-12cm}  


\begin{thebibliography}{99}
\bibitem{wang} F. Wang and K. Hauser, "Dense robotic packing of irregular and novel 3D objects," IEEE Transactions on Robotics 38, No.2, pp. 1160-1173, 2021.

\bibitem{dong} D. Siyuan and A. Rodriguez, "Tactile-based insertion for dense box-packing," in 2019 IEEE/RSJ International Conference on Intelligent Robots and Systems (IROS), 2019.

\bibitem{ts-nvae} R. Okumura, N. Nishio, and T. Taniguchi, "Tactile-sensitive NewtonianVAE for high-accuracy industrial connector-socket insertion," in International Conference on Intelligent Robots and Systems (IROS), 2022.

\bibitem{world_model} D. Ha and J. Schmidhuber, "Recurrent world models facilitate policy evolution," in NeurIPS, 2018.

\bibitem{nvae} M. Jaques, M. Burke, and T. Hospedales. "NewtonianVAE: Proportional control and goal identification from pixels via physical latent spaces," in 2021 IEEE/CVF Conference on Computer Vision and Pattern Recognition (CVPR), pp. 4452-4461, 2021.
\bibitem{lee} A. X. Lee, A. Nagabandi, P. Abbeel, and S. Levine, "Stochastic latent actor-critic: Deep reinforcement learning with a latent variable model," in arXiv, 2019.

\bibitem{levine} S. Levine, "Reinforcement learning and control as probabilistic inference: Tutorial and review," in arXiv, 2018.

\bibitem{hafner} D. Hafner, T. Lillicrap, J. Ba, and M. Norouzi, "Dream to control: Learning behaviors by latent imagination," in ICLR, 2020.

\bibitem{dreaming} M. Okada and T. Taniguchi, "Dreaming: Model-based reinforcement learning by latent imagination without reconstruction," in ICRA, pp. 4209-4215, 2021.

\bibitem{dreamingv2} M. Okada and T. Taniguchi, "Dreamingv2: Reinforcement learning with discrete world models without reconstruction," in International Conference on Intelligent Robots and Systems (IROS), 2022.

\bibitem{multi-dreaming} A. Kinose, M. Okada, R. Okumura, and T. Taniguchi, "Multiview dreaming: Multi-view world model with contrastive learning," in arXiv, 2022.

\bibitem{hafner2} D. Hafner, T. Lillicrap, I. Fischer, R. Villegas, D. Ha, H. Lee, and J. Davidson, "Learning latent dynamics for planning from pixels," in ICML, 2019.

\bibitem{okada} M. Okada, N. Kosaka, and T. Taniguchi, "Planet of the bayesians: Reconsidering and improving deep planning network by incorporating bayesian inference," in International Conference on Intelligent Robots and Systems (IROS), 2020.

\bibitem{daac} R. Okumura, M. Okada, and T. Taniguchi, "Domain adversarial and conditional state space model for imitation learning," in IROS, pp. 5179-5186, 2020.

\bibitem{levine2} S. Levine and P. Abbeel, "Learning neural network policies with guided policy search under unknown dynamics," in NeurIPS, 2014.
\bibitem{levine3} S. Levine, N. Wagener, and P. Abbeel, "Learning contact-rich manipulation skills with guided policy search," in ICRA, 2015.
\bibitem{watter} M. Watter, J. Springenberg, J. Boedecker, and M. Riedmiller, "Embed to control: A locally linear latent dynamics model for control from raw images," in NeurIPS, 2015.

\bibitem{vae} D. P. Kingma and M. Welling, "Auto-encoding variational bayes," in ICLR 2014.

\bibitem{leakyrelu} A. L. Maas, A. Y. Hannun, and A. Y. Ng. "Rectifier nonlinearities improve neural network acoustic models," in Proc. ICML, Vol. 30, No. 1, 2013.

\bibitem{cfil} E. Johns, "Coarse-to-fine imitation learning: Robot manipulation from a single demonstration," in ICRA, pp. 4613-4619, 2021.





\bibitem{adam} D. Kingma and J. Ba, "Adam: A method for stochastic optimization," in ICLR, 2014.



\end{thebibliography}
\end{document}